\documentclass{article}

\PassOptionsToPackage{numbers, compress}{natbib}

\usepackage[preprint]{neurips_2024}

\usepackage[utf8]{inputenc} 
\usepackage[T1]{fontenc}    
\usepackage{hyperref}       
\usepackage{url}            
\usepackage{booktabs}       
\usepackage{amsfonts}       

\usepackage{microtype}      
\usepackage[dvipsnames]{xcolor}         
\usepackage{enumerate}

\usepackage{graphicx}
\usepackage{booktabs}

\usepackage{amsmath,amssymb,amsthm}
\usepackage{array}

\usepackage{algorithm,algorithmicx,algpseudocode}
\usepackage{caption}
\usepackage{comment}
\usepackage[nice]{nicefrac}

\usepackage{xspace}
\usepackage{colortbl}

\usepackage{enumitem}
\usepackage{subcaption}
\usepackage{epigraph}
\usepackage{changepage}
\usepackage{multirow}

\definecolor{citecolor}{HTML}{0071BC}
\definecolor{linkcolor}{HTML}{ED1C24}

\newcommand{\codecomment}[1]{{\color{gray}{#1}}}
\newlength\savewidth\newcommand\shline{\noalign{\global\savewidth\arrayrulewidth
  \global\arrayrulewidth 1pt}\hline\noalign{\global\arrayrulewidth\savewidth}}
\newcommand{\tablestyle}[2]{\setlength{\tabcolsep}{#1}\renewcommand{\arraystretch}{#2}\centering\footnotesize}
\definecolor{baselinecolor}{HTML}{d6eaf8}
\newcommand{\baseline}[1]{\cellcolor{baselinecolor}{#1}}

\newcolumntype{x}[1]{>{\centering\arraybackslash}p{#1pt}}
\newcolumntype{y}[1]{>{\raggedright\arraybackslash}p{#1pt}}
\newcolumntype{z}[1]{>{\raggedleft\arraybackslash}p{#1pt}}

\def \etal{{\em et al.}}

\title{Improving Joint Embedding Predictive Architecture with Diffusion Noise}

\author{
Yuping Qiu$^{1}$\thanks{Optional footnote for author info (e.g., webpage).}, 
Rui Zhu$^{2}$, 
Ying-cong Chen$^{1}$ \\
$^{1}$The Hong Kong University of Science and Technology (Guangzhou) \\
$^{2}$The Chinese University of Hong Kong, Shenzhen \\
\vspace{0.6em} 
}

\begin{document}

\maketitle

\begin{abstract}

Self-supervised learning has become an incredibly successful method for feature learning, widely applied to many downstream tasks. It has proven especially effective for discriminative tasks, surpassing the trending generative models. However, generative models perform better in image generation and detail enhancement. Thus, it is natural for us to find a connection between SSL and generative models to further enhance the representation capacity of SSL. As generative models can create new samples by approximating the data distribution, such modeling should also lead to a semantic understanding of the raw visual data, which is necessary for recognition tasks. This enlightens us to combine the core principle of the diffusion model: diffusion noise, with SSL to learn a competitive recognition model. Specifically, diffusion noise can be viewed as a particular state of mask that reveals a close relationship between masked image modeling (MIM) and diffusion models. In this paper, we propose N-JEPA (Noise-based JEPA) to incorporate diffusion noise into MIM by the position embedding of masked tokens. The multi-level noise schedule is a series of feature augmentations to further enhance the robustness of our model. We perform a comprehensive study to confirm its effectiveness in the classification of downstream tasks. Codes will be released soon in public.

\end{abstract}

\section{Introduction}
\label{sec:intro}



\begin{figure*}[t!]
\centering
\includegraphics[width=\linewidth]{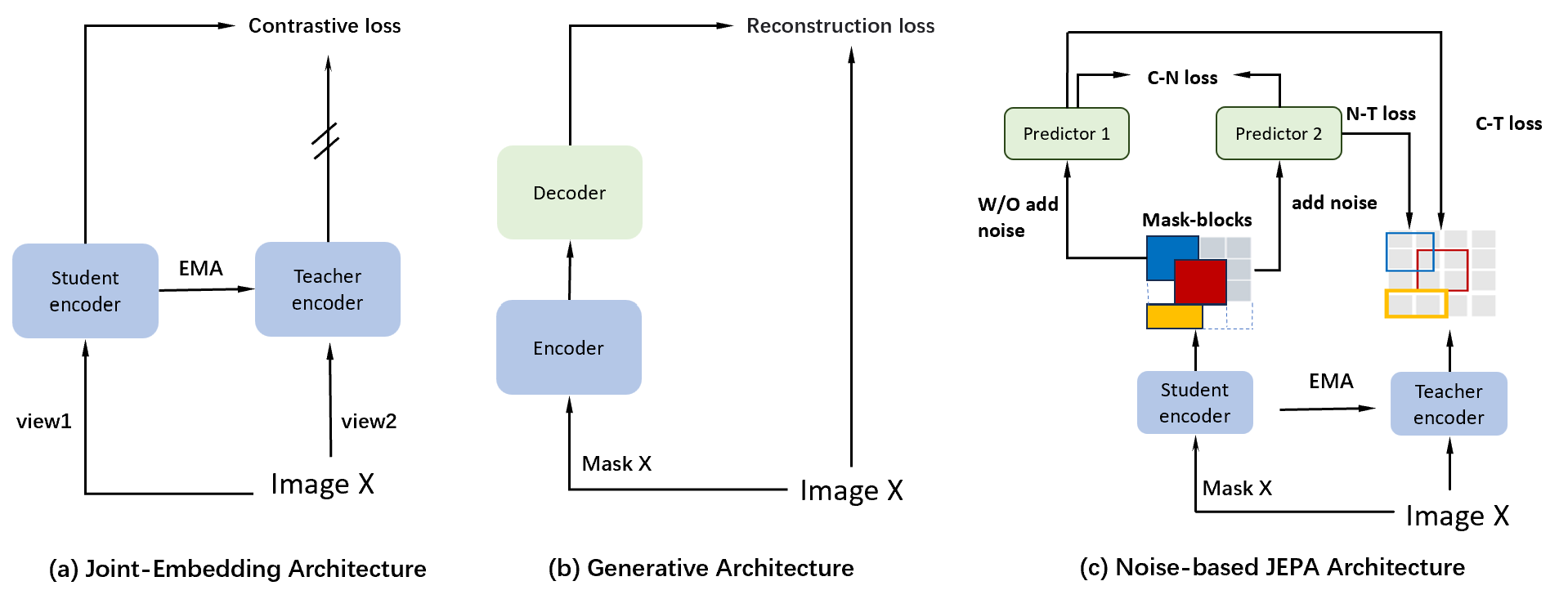}
\caption{\small \textbf{ Main types of self-supervised learning}. (a) Contrastive Learning: The augmented two views are fed into the student and teacher encoder, and the contrastive loss aims at pulling two feature embeddings closer. (b) MIM: The reconstructive loss aims at recovering pixel or feature-level tokens with the input image as the label. (c) N-JEPA: Our goal is to predict the teacher feature from the noised feature in the encoder space, 
with the multi-level noise schedule being the key difference between the two predictors.
Predictor 1 focuses on the context features of masked blocks, while predictor 2 predicts from the noised features. 
More details in Section~\ref{sec:method}.
}
\label{fig:figure_intro_2}
\end{figure*}

Recent years have witnessed the success of Self-supervised learning (SSL), which utilizes unlabeled data to achieve high-quality feature representations by solving proxy tasks and the corresponding pseudo-labels. Such as contrastive learning (Figure \ref{fig:figure_intro_2}a)~\cite{caron2021unsupervised,he2020momentum,chen2020improved} heavily relying on data augmentation invariance, and Mask Image Modeling (MIM in Figure \ref{fig:figure_intro_2}b)~\cite{zhou2021ibot,grill2020bootstrap,wei2022masked,xie2022simmim} predicting masked pixels or tokens given visual contents.
However, the hand-crafted data augmentations are limited to human prior, which can not easily generalize on other modalities~\cite{assran2023selfsupervised,verma2021domainagnostic}. MIM could alleviate such problems while low-level representations~\cite{tao2022siamese} hinder performance in off-the-shelf evaluations (e.g., linear-probing) or transfer settings with limited supervision for classification tasks.

Notably, the domination of denoising diffusion models~\cite{luo2023latent,ho2020denoising,song2020denoising} in image generation has gradually affected the development of SSL. 
On the one hand, \cite{li2023dreamteacher,xiang2023denoising} propose that generative diffusion models can be leveraged as a strong pre-trained representation for downstream tasks. However, the performance still lags behind the semantic representations of SSL.
On the other hand, DiffMAE~\cite{wei2023diffusion} first establishes the connection between diffusion models and SSL, suggesting that MAE~\cite{he2022masked} can be viewed as a single-step, patch-conditioned diffusion model.
Besides,  the methodology of diffusion model is adding noise and then denoising, which is consistent with the philosophy of SSL: corruption first and then reconstruction~\cite{balestriero2023cookbook,Liu_2021,Ericsson_2022}.
These observations inspire us to inject the diffusion noise to enhance the pre-training process of SSL and achieve good representations.
 
We propose N-JEPA to improve the pre-training process of the Joint-Embedding Predictive Architecture (JEPA])~\cite{assran2023selfsupervised} by predicting the target feature from the noised feature in the encoder space. 
Our approach is straightforward in that we introduce EDM noise to the position embedding of the masked tokens in the representation space of JEPA. EDM~\cite{karras2022elucidating} utilizes the modified distribution $P_\sigma(x)$ rather than $P_t(x)$, and $P_\sigma(x) = P_{data}(x) * \mathcal{N}(0, \mathbf{\sigma^2I}) $, thus we avoid the need to change the ViT framework to incorporate timestep embedding. (See Figure \ref{fig:figure_intro_2}(c)). 
In addition, we initialize different mask blocks with various sampling noises from the same noise distribution. This approach has two advantages: 
Firstly, adding noise to the position embedding of the mask blocks provides a disturbance to the deterministic positions, so our model can learn more diverse features. 
Secondly, a multi-level noise schedule can be seen as a form of feature-level augmentations~\cite{li2023metaug,wu2023hallucination}, further enhancing the model's robustness without relying on hand-crafted augmentations and simultaneously avoiding introducing strong bias. Our contributions are as follows:
\begin{itemize}[]
    \item We propose N-JEPA  to build a connection between SSL and diffusion models.  
    \item The proposed multi-level noise schedule can be viewed as a kind of feature augmentation that could further improve the robustness of our model.
    \item We compare N-JEPA with previous baselines and the results are substantially better than the off-the-shelf counterparts in classification downstream tasks. Our comprehensive empirical studies confirm N-JEPA's effectiveness.
\end{itemize}

\section{Related work}
\subsection{Mask Image Modeling}

Inspired by BERT~\cite{devlin2018bert}, which predicts text tokens from masked tokens, BEiT~\cite{bao2021beit} first proposed Masked Image Modeling for self-supervised visual learning. However, BEiT relies on a pre-trained autoencoder~\cite{bao2021beit} to get discrete visual tokens, which is time-consuming. 
So MAE~\cite{he2022masked} simplifies the training pipeline by applying random masks to the input image patches and directly reconstructing the masked image patches. 
Furthermore, CrossMAE~\cite{fu2024rethinking} delves into studying the mask strategies and proposes that random masking is ineffective due to the highly redundant information in image data. For efficiency, the decoder of CrossMAE only leverages cross-attention between masked and visible tokens. AttMask~\cite{kakogeorgiou2022hide} focuses on the masked tokens from the attention map to create a more challenging MIM task.
While Adam \etal~\cite{fu2024rethinking} argue that self-attention is not essential for good representation learning.
Recently, Yann LeCun~\etal~\cite{assran2023selfsupervised} introduced the Image-based Joint-Embedding Predictive Architecture (I-JEPA), which aims at predicting to map the masked patches within a high-level representation space. I-JEPA allows the model to concentrate more on semantic features, enhancing the ability to understand and predict across different modalities.
Based on I-JEPA, FlexPredict~\cite{bar2024stochastic} introduces noise to tackle location uncertainty. However, FlexPredict needs to learn an extra elaborated matrix $A$, which forces the model to balance the location certainty and the influence of context features in predictions. Thus, it can prevent the stochastic positional embedding from collapsing into the deterministic one.


%

\subsection{Diffusion model} 
Denoising diffusion probabilistic models (DDPMs)~\cite{ho2020denoising} have emerged as the leading paradigm in generative models owing to the exceptional capability to produce high-quality samples~\cite{rombach2022highresolution,peebles2023scalable} and the proficiency in synthesizing intricate visual concepts~\cite{nichol2022glide,ramesh2022hierarchical}.
The basic idea of diffusion models works with continuous or discrete noise injection on  data~\cite{mittal2022points}  or latent space (latent diffusion model~\cite{rombach2022highresolution}) and learning the reverse denoising process. However, the development of DDPM is blocked by its inherent limitations, such as the slow sampling speed and the heavy training cost. Therefore, some works focus on accelerating sampling, including Discretization Optimization~\cite{ruemelin1982numerical,song2020score}, Non-Markovian Process~\cite{song2020denoising}, and Partial Sampling~\cite{salimans2022progressive,meng2023distillation,luo2023latent,zhao2023mobilediffusion}. In particular, Diffusion distillation~\cite{salimans2022progressive} is a highly effective method for reducing the number of sampling steps in a diffusion model through step distillation. 
Additionally, some approximate maximum likelihood training~\cite{nichol2021improved,song2021maximum} and training loss weighting methods~\cite{karras2022elucidating,kim2021soft} have been proposed to improve the training efficiency of diffusion models. To further enhance the scalability of the diffusion model, recent works proposed various Transformer-based  architecture~\cite{peebles2023scalable,bao2022all,gao2023masked,zheng2023fast}. For instance, GenViT~\cite{yang2022your} has shown that ViT has inferior performance compared to UNet on generation tasks. In comparison, U-ViT~\cite{bao2022all} achieves competitive performance with a UNet-like network by adding long-skip connections and convolutional layers.



\subsection{Combination between SSL and Diffusion models}

A natural idea is to combine SSL and diffusion models to enhance the performance of each other. For example, MaskDit~\cite{zheng2023fast} and MDT~\cite{gao2023masked} leverage the masking paradigm of SSL to improve diffusion models' training efficiency significantly. MDT~\etal ~\cite{gao2023masked} introduces a mask latent scheme to explicitly enhance the ability of diffusion models for contextual relation learning among object semantic parts in an image. MaskDiT~\cite{zheng2023fast} proposes the fast training with masked Diffusion Transformers~\cite{peebles2023scalable} by introducing an asymmetric encoder-decoder architecture and a new training objective. Similarly, recent works proposed utilizing diffusion noise to boost the pretraining of SSL. DiffMAE~\cite{wei2023diffusion} links diffusion noise with MAE~\cite{he2022masked} as a single-step patch-conditioned diffusion model. DreamTeacher~\cite{li2023dreamteacher} suggests distilling knowledge from well-trained generative models into standard image backbones because the high-quality samples generated by the generative models can guide the model to learn the internal representation of the data. Recently, IWM~\cite{garrido2024learning} further leverages I-JEPA to learn an Image World Model (IWM) and shows that it relies on three key aspects: conditioning, prediction difficulty, and capacity. DDAE~\cite{xiang2023denoising}confirms that denoising diffusion autoencoders can learn strongly linear-separable feature representations in the middle of up-sampling and highlights the underlying nature of diffusion models as unified self-supervised learners. L-DAE ~\cite{chen2024deconstructing} deconstructs a DDM and transforms it into a classical Denoising Autoencoder to explore the critical modern components for self-supervised representation learning.

\begin{figure*}[ht!]
\centering
\includegraphics[width=1\linewidth]{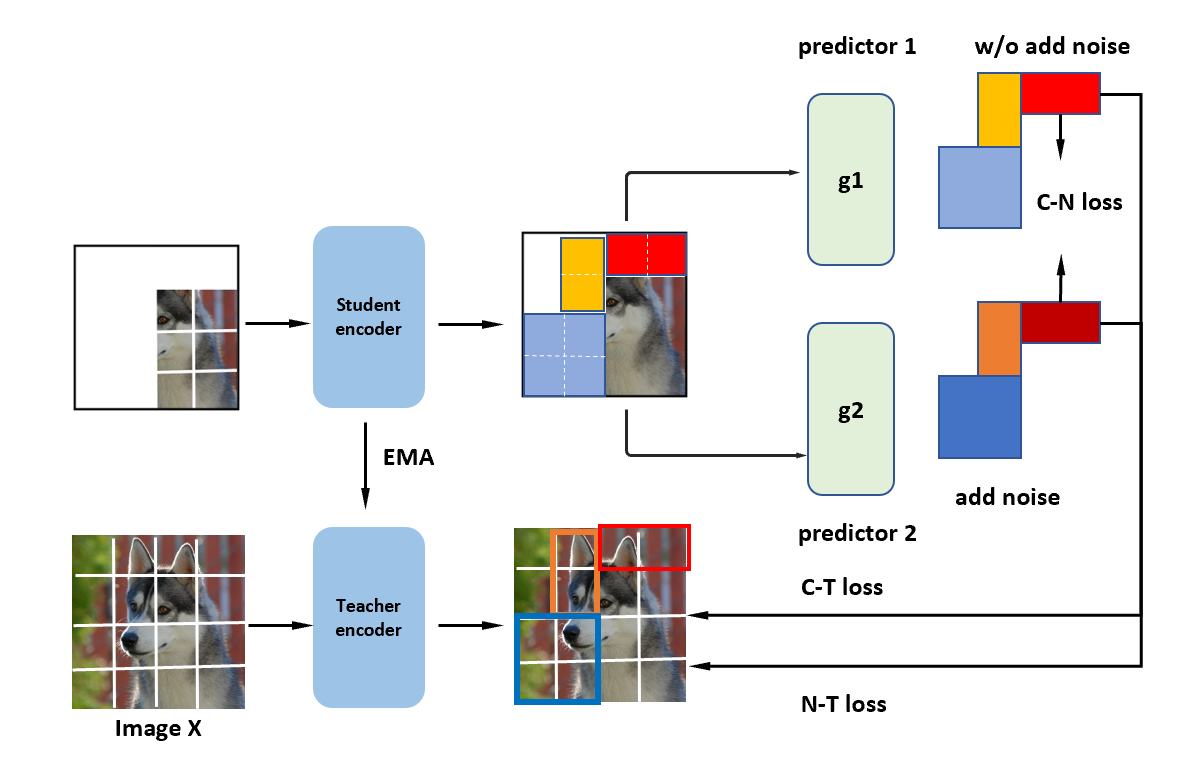}
\caption{\small \textbf{The overview of our N-JEPA}. The image $X$ will be converted into a sequence of $N$ non-overlapping patches and fed into the teacher encoder, and we feed only visible patches to the student encoder. We aim to predict the representations of various masked blocks shown in different colors (red, yellow, blue). Whether adding mutli-level noise schedule is the difference between two predictors, other settings are the same. The darker colors mean that we have already added noise to masked blocks. $C-T$ loss means context-teacher loss, we do not add noise on mask position embedding, so predictor $1$ obtains the context representations, $N-T$ means noise-teacher, and we have noisy representations from predictor $2$. $C-N$ is the denoise loss, which does the denoising process between context and noisy features. }
\label{fig:method_full}
\end{figure*}

\section{Method}
\label{sec:method}
\subsection{Preliminary}
DDPMs~\cite{ho2020denoising} or DDIMs~\cite{song2020denoising} are training with a forward noising process and a reverse denoising process. In the forward process, we gradually introduce Gaussian noise $\epsilon$ to the data distribution $P_{data}(x)$, thereby obtaining a series of noise-perturbed latent variables of the original samples ($x_{1}$, $x_{2}$, ..., $x_{T}$ ). If the timestep $T$ is large enough, $x_{T}$ would be an isotropic Gaussian noise. 
The forward process can be defined as $q(x_t | x_0) = \mathcal{N}(\alpha_t x_0, \sigma_t^2 \mathbf{I})$, where $\alpha_t$ and $\sigma_t$ are hyper-parameters that control the signal-to-noise ratio. 
Similarly, the objective of the reverse process is to predict the noise introduced by the forward process at each timestep, originating from $x_{T}$, and gradually remove the noise to generate new samples that align with the original data distribution $P_{data}(x)$. 
We define the reverse process with learnable Gaussian transitions parameterized by $\theta$: $p_\theta(x_{t-1} | x_t) = \mathcal{N}(\mu_\theta(x_t, t), \Sigma_t^2 \mathbf{I})$, where mean $\mu_\theta(x_t, t)$ is predicted by networks and variance $\Sigma_t^2$ is a constant value. Furthermore, Song\etal~\cite{song2020score} proposes score SDE, which is a unified continuous framework based on the stochastic differential equation to describe DDPMs~\cite{ho2020denoising} and NCSN~\cite{song2019generative}.
\begin{align}
    dx = f(x,t)dt + g(t)dw, \label{con:forward SDE}
\end{align}
Equation~\ref{con:forward SDE} shows the forward SDE, the function $f(\cdot)$ and $g(t)$ are referred to as the drift coefficient and the diffusion coefficient. $w$ is a standard Brownian motion, and $dw$ can be viewed as white noise. Unlike the forward SDE process, the reversed SDE process is defined in terms of the reverse-time Stochastic Differential Equation \ref{con:reverse SDE}, by operating in a backward time manner:
\begin{align}
    dx = [f(x,t)-g(t)^2 \nabla_{x}logp_{t}(x)]dt + g(t)d\overline{w}. \label{con:reverse SDE}
\end{align}
In DDPMs, $\alpha_t = \sqrt{\Pi_{i=1}^{t}{(1-\beta_i)}}$ and $\alpha_t^2 + \sigma_t^2 = 1$. $\beta_{1...T}$ are sampled by a linear schedule from $\beta_{min}$ to $\beta_{max}$. However, instead of using Variance Preserving parameterization, EDM~\cite{karras2022elucidating} chooses to use the "Variance Exploding" parameterization where we add Gaussian noise with $ \mathcal{N}(0, \mathbf{\sigma^2I})$ into the data distribution. To be specific, EDM~\cite{karras2022elucidating} utilizes modified distribution $P_\sigma(x)$ rather than $P_t(x)$, and $P_\sigma(x) = P_{data}(x) * \mathcal{N}(0, \mathbf{\sigma^2I}) $, where $*$ denotes the convolution operation. So the diffused data $x_\sigma$ can be formulated as:
\begin{align}
    x_\sigma = x_0 + n, n \sim \mathcal{N}(0, \mathbf{\sigma^2I}), \label{con:add EDM noise}
\end{align}
where $x_0$ belongs to $P_{data}(x)$. Without scaling, Equation \ref{con:reverse SDE} can be simplified as:
\begin{align}
    dx = -\sigma \nabla_{x}logp_\sigma(x)d\sigma,   \sigma \in [\sigma_{min},\sigma_{max}].\label{con:EDM SDE}
\end{align}
In this way, we can use score-based SDE to unify diffusion models, as $\nabla_{x}logp_\sigma(x)$ is the score function.
Ideally, we hope to select $\sigma(t)$ in such a way that $P_{\sigma_{min}} \approx P_{\sigma_{data}}, P_{\sigma_{max}} \approx \mathcal{N}(0, \mathbf{\sigma^2_{max}I})$.  
In practice, if $\sigma_{max} \gg \sigma_{data} $, we can consider $P(x; \sigma_{max})$ to be a pure Gaussian noise with a variance close to $\sigma_{max}$. Unlike the previous methods~\cite{song2020denoising,ho2020denoising}, EDM considers directly estimating the denoising function of the denoised samples $D(x;\sigma)$:
\begin{align}
    \mathbb{E}_{x_0 \sim \mathbf{p_{data}}}\mathbb{E}_{n \sim \mathcal{N}(0, \mathbf{\sigma^2I})} \parallel D_\theta(x_0 +n;\sigma) - x_0 \parallel ^2_{2},\\
    \nabla_{x}logp_\sigma(x) = (D_\theta(x_0 +n;\sigma) - x_\sigma) / {\sigma^2}.
\end{align}
where $x_0$ represents the training sample, and $n$ is the added noise. In this scenario, the calculation of the score function has transformed into estimating $D(x;\sigma)$ for the added noise.\\

In this paper, our method N-JEPA follows the diffusion noising schedule of EDM~\cite{karras2022elucidating} for two reasons: 1) The design of EDM puts diffusion models into a common framework which enables it to be compatible with various earlier diffusion models~\cite{ho2020denoising,song2020denoising}. 2) Since EDM utilizes $P\sigma(x)$ instead of $P_t(x)$, we can preserve the ViT framework rather than introducing extra $t$ embeddings to a large extent, which will not bring extra computational cost.

\subsection{Overall Architecture}
In this study, we explore the effectiveness of injecting diffusion noise into JEPA~\cite{assran2023selfsupervised} to enhance the pretraining process of SSL. I-JEPA has already emphasized the importance of acquiring semantic understanding in self-supervised representations without relying on additional prior knowledge encoded through image transformations. In this way, our model will investigate how to ingeniously combine diffusion noise with JEPA architecture to release the power of SSL. The overall architecture of N-JEPA is illustrated in Fig.\ref{fig:method_full}. 

\textbf{Joint-Embedding Predictive Architecture.} First, let us review the difference between joint-embedding-predictive-architecture and the generative method. Generative methods attempt to directly reconstruct the missing information from input x, using a decoder network conditioned on latent variables to aid the reconstruction process. In comparison, the joint-embedding-predictive architecture uses a predictor network to facilitate the prediction process. Instead of predicting the input space, we predict the representation space to get high-level semantic representations. To be specific, JEPA uses a predictor network that is conditioned on position embeddings corresponding to the location of the target block in the image. Moreover, in generative models, the decoder is typically a lightweight  Vision Transformer (ViT), while the predictor is responsible for predicting features, so it generally adopts a ViT structure similar to the encoder but with a slightly smaller depth.

\textbf{Input.} During training, the image $X$ will be converted into a sequence of $N$ non-overlapping patches and fed into the teacher encoder to get the corresponding patch-level representation $z_t=\{z_{t_1}, \dots, z_{t_N}\}$ where $z_{t_k}$ is the representation associated with the $k^\text{th}$ patch. We also denote by $z_s=\{z_{s_1}, \dots, z_{s_N}\}$ the corresponding patch-level representation obtained by student encoder. The parameters of the student encoder network are Exponentially Moving Averaged (EMA) to the parameters of the teacher encoder network. To better illustrate our objective, we randomly select $L$ blocks from $z_t$ to apply masking. So $z_t(i)=\{z_{t_{j}}\}_{j \in L_i}$ is the corresponding patch-level representation, the same as $z_s=\{z_{s_{j}}\}_{j \in L_j}$ where $L_j$ the mask associated with the visible patches $j$. In our experiments, we set $L$ to 4 and randomly sample the blocks with an aspect ratio ranging from 0.75 to 1.5 and a scale in the range of (0.15, 0.2). In section \ref{sec:experiment}, we will provide a detailed explanation of the masking strategy.

\textbf{Predictor and Loss.} Two narrow Vision Transformer (ViT) predictors utilize the student encoder output as their input. Conditioned on positional visible tokens, they predict the corresponding representations of teacher blocks, as indicated by the colored boxes at the teacher branch. The only difference between the two predictor networks is the EDM noise. To simplify, predictor $1$ does not add noise by default, so it will predict the corresponding representations without adding noise on position embeddings. While predictor $2$ does, with different initialized noise added to masked blocks. Multi-level noise schedule aims to initialize $L$ times of different noise for $L$ mask blocks. All noise follows the same distribution. Based on our objectives, our losses can be divided into two types:
\textbf{prediction loss} and \textbf{denoise loss}. The former involves predicting the features of the corresponding block in the teacher branch through different predictors, for which we employ a simple smooth-L1 loss. Smooth-L1 loss is a smooth version of L1 Loss, which can solve the problem of gradient explosion caused by outliers, making the training process more stable. The latter is about denoising the output from the two predictors, for which we utilize the MSE loss. i.e.,

\textit{\textbf{Prediction loss.}} Where $\hat{z}_t(i)$ is the representation of teacher block.$z_{s_c}(i)$ is the predicted representation by predictor $1$,  $z_{s_N}(i)$ by predictor $2$.

\begin{align}
 L_{C-T} =\frac{1}{L} \sum^L_{i=1} {D}\left(\hat{z}_t(i), z_{s_c}(i)\right) = \frac{1}{L} \sum^L_{i=1} \begin{cases}
0.5 * \sum_{j \in L_i} \lVert\hat{z}_{t_j} - z_{s_cj}\rVert^2_2 , & if \lVert\hat{z}_{t_j} - z_{s_cj}\rVert<1 \\
\lVert\hat{z}_{t_j} - z_{s_cj}\rVert - 0.5,& otherwise \\
\end{cases}
\end{align}
N-T loss means the loss between the noisy representations and the representations of the teacher blocks.
\begin{align}
 L_{N-T} =\frac{1}{L} \sum^L_{i=1} {D}\left(\hat{z}_t(i), z_{s_N}(i)\right) = \frac{1}{L} \sum^L_{i=1} \begin{cases}
0.5 * \sum_{j \in L_i} \lVert\hat{z}_{t_j} - z_{s_Nj}\rVert^2_2 , & if \lVert\hat{z}_{t_j} - z_{s_Nj}\rVert<1 \\
\lVert\hat{z}_{t_j} - z_{s_Nj}\rVert - 0.5,& otherwise \\
\end{cases}
\end{align}

\textit{\textbf{Denoise loss.}} The loss is simply the average $L2$ distance between the context representations and noisy representations.
\begin{align}
    L_{C-N} =\frac{1}{L} \sum^L_{i=1} {D}\left({z}_{s_N}(i), z_{s_C}(i)\right) = \frac{1}{L}  \sum^L_{i=1}\sum_{j \in L_i} \lVert{z}_{s_Nj} - z_{s_cj}\rVert^2_2. 
\end{align}

\textit{\textbf{Overall loss.}} $\lambda_{1}$, $\lambda_{2}$ are the hyper-parameters, in section \ref{sec:experiment}, we find that giving them a small value will help training.
\begin{align}
    L_{total} =  L_{C-T} + \lambda_{1}L_{N-T} + \lambda_{2} L_{C-N}.
\end{align}

\section{Experiments}
\label{sec:experiment}
\subsection{Implementation Details}
In this section, we will provide a detailed description of the model architecture, masking strategy, training setup, and evaluation settings. Our model is pre-trained on the ImageNet-1K(IN-1K) training set for 100/600 epochs. During the training process, we observed an intriguing phenomenon where the performance of linear probing using the weights trained for 80 or 550 epochs was better than that of the weights trained for 100 or 600 epochs. This contradicts common expectations and prompts us to investigate the cause of this discrepancy. Upon investigation, we discovered that all hyper-parameter schedules were scaled 25\% beyond the actual training schedule.(see Figure~\ref{fig:grad_stats} in Appendix) This is due to the last 25\% of the default scheduler period making hyper-parameter updates too aggressive. By simply truncating the schedulers, we set $ipe_{scale} = 1.25$ to ensure a fair comparison when training for 600 epochs. We evaluate the linear-probing performance on ImageNet-1K using both 100\% and only 1\% , 10\% of the available labels to demonstrate whether N-JEPA has acquired high-semantic and robust representations without relying on hand-crafted data augmentations.
\subsection{Model Architecture} The overall architecture is based on Vision Transformers (ViT) to ensure compatibility with the most widely used SSL frameworks. Following the setting of I-JEPA~\cite{assran2023selfsupervised}, we use a ViT architecture for the student-encoder, teacher-encoder, and predictor networks. Considering the computational resources, we limit our experiments by utilizing ViT-Base for the ViTs, excluding larger-scale models such as ViT-Huge and ViT-Giant.  During pretraining, the student-encoder and teacher-encoder are vanilla ViT-Base of depth 12 and width 768 without any modification. For the predictor, we set the depth of the predictor to 6. So our predictors are based on the same architecture with a smaller depth and fixed embedding dimension of 384. The teacher encoder is the EMA of the student encoder, and the momentum coefficient increases from 0.996 to 1.0 at the end of training. For the multi-noise schedule, we follow the default parameters of EDM ($P_{mean}$ = -1.2, $P_{std}$ = 1.2, $\sigma_{data}$ = 0.5). More pretraining settings can be seen in Appendix~\ref{sec:implementation details}.

\subsection{Masking Strategy} We adopt the multi-block masking approach from the pretraining method I-JEPA~\cite{assran2023selfsupervised} as it is crucial in acquiring more semantic representations than traditional block and random masking strategies. Specifically, as a default setting, a mask $L_x$ consisting of 4 teacher block masks is sampled, with random scales ranging between 0.15 and 0.2 and aspect ratio within (0.75, 1.5), allowing for the possibility of overlap. Additionally, we sample one student block mask with a random scale in the range of (0.85, 1.0) and a unit aspect ratio. Subsequently, we remove any regions in the context block mask that overlap with any of the four teacher block masks. It is important to note that the student and teacher block masks are sampled independently for each image in the mini-batch.

\subsection{Training Setup}
We conduct all the experiments on ImageNet-1K with 224×224 resolution and a batch size of 128 for 100 epochs and 1024 for 600 epochs due to the limitation of computational resources. We believe that a larger batch size could result in more performance gains. We use AdamW to optimize the student encoder and predictor weights. The learning rate is linearly increased from $10^{-4}$ to $10^{-3}$ during the first 40 epochs of pretraining, then becomes a constant $1e^{-3}$
learning rate. The cosine weight decay schedule goes from 0.04 to 0.4 during pretraining. All models for 100 epochs are trained with 8 RTX 3090
nodes and 8 NVIDIA V100 nodes for 600 epochs.
\subsection{Linear evaluation}
We report results on the image classification tasks using linear probing to demonstrate that N-JEPA learns robust representations. In this section, self-supervised models are pre-trained on the ImageNet-1K dataset. The pre-trained model weights are then frozen, and a linear classifier is trained using the full ImageNet-1K training set. During the evaluation phase, we employ the student encoder to learn and create a comprehensive global image representation by averaging its output, moving away from reliance on the [cls] token. Following DINO\cite{caron2021emerging}, the linear classifier undergoes training using SGD with a batch size of 1024 for 50 test epochs on the ImageNet-1K dataset. Throughout the linear evaluation, as shown in table~\ref{tb:loss_weight}, we conduct an exploration of various weight settings and table~\ref{tb:lineareval} provides a comprehensive report on the Top-1 accuracy.

\begin{table}[t]
    \centering
    \begin{tabular}{l| l l c }
        \bf\small Method & \bf\small Arch. & \bf\small  Epochs & \bf\small Top-1\\
        \shline
        
        \small Baseline (C-T) & \small ViT-B/16 & 100 & 66.8\\[1ex]
        \small(C,T) + (N,C)  & \small ViT-B/16 & 100 & 66.4\\[1ex]
        \small(C,T) + (N,T)  & \small ViT-B/16 & 100 & 65.3\\[1ex]
        \small \baseline{(C,T) + (N,T) + (N,C)} & \small \baseline{ViT-B/16} & \baseline{100} & \baseline{67.2}\\[1ex]
       
    \end{tabular}
    \vspace{0.5em}
    \caption{{\bf Loss selection}. Linear evaluation on different loss choices. I-JEPA uses context prediction loss (C-T). In this table, we only analyze loss choices without considering the weight of each loss. So all weights are equal to $1$. We observe that only (N-C) or (N-T) will harm the performance.}
  \label{tb:loss_choice}
  
\end{table}

\textbf{Loss selection.} From table \ref{tb:loss_choice}, by adding only N-T loss or C-N loss, we observe a slight decrease in the performance of linear probing. However, when our total loss incorporates noisy prediction loss (N-T loss) and denoise loss without modifying the weights, the performance improves by 0.4\%, indicating the effectiveness of our loss function.

\textbf{Loss weight.} Table \ref{tb:loss_weight} shows the performance of our total loss under different weights. We find that no matter what we train for 100 or 600 epochs, the results demonstrate that the weights of noisy prediction loss and denoise loss should be relatively low. This aligns with our understanding of the significance of context prediction loss, with the other losses serving as auxiliary components to further enhance the robustness of our model.

\begin{table}[!ht]
    \centering
    \begin{tabular}{l| l l c }
        \bf\small Method & \bf\small Arch. & \bf\small  Epochs & \bf\small Top-1\\
        \shline
        \multirow{2}{*}{\small total loss (CT 1 NT 1 NC 1) }
         & \small ViT-B/16 & 100 & 67.2\\
         & \small ViT-B/16 & 600 & 72.5\\[1ex]
         \multirow{2}{*}{\small total loss ( CT 1, NT 1, NC 0.1) }  & \small ViT-B/16 & 100 & 64.0\\
         & \small ViT-B/16 & 600 & 71.6\\[1ex]
         \multirow{2}{*}{\small  total loss ( CT 1, NT 0.1, NC 1) }  & \small ViT-B/16 & 100 & 64.8\\
         & \small ViT-B/16 & 600 & 71.8\\[1ex]
        \multirow{2}{*}{\small \bf{total loss ( CT 1, NT 0.1, NC 0.1)}} & \small \baseline{ViT-B/16} &  \baseline{100} & \baseline{\bf 67.9}\\ 
        & \small \baseline{ViT-B/16} & \baseline{600}  & \baseline{\bf 73.4}
    \end{tabular}
        \vspace{0.5em}
    \caption{{\bf The weights of loss hyper-parameters}. We see a large performance improvement with lower weights of N-T loss and N-C loss. (+1.1\%, +1.3\% for 100 and 600 epochs VS. baseline.) }
  \label{tb:loss_weight}
\end{table}

\textbf{Multi-level Noise schedule.} The task of the diffusion model is to gradually transform a noise input into a high-quality and diverse image. The original noise schedule introduces timestep embedding, while in our framework, we implicitly avoid adding $t$ by introducing EDM. EDM utilizes modified distribution $P_\sigma(x)$ instead of $P_t(x)$ as seen in Preliminaries. Regarding noise schedules, there are two options: single-level and multi-level. Single-level noise means that we initialize noise once and add it to the position embeddings of mask blocks, while multi-level noise schedule aims to initialize $L$ times of different noise for $L$ mask blocks. All noise follows the same distribution. We compared the single-level of noise on position embedding in different mask blocks, and the table \ref{tb:add_multi_noise} shows that a multi-level noise schedule further enhances the linear evaluation performance.

\begin{table}[ht]
    \centering
    \begin{tabular}{l| l l c }
        \bf\small Exps  & \bf\small  Epochs & \bf\small  Noise type & \bf\small Top-1\\
        \shline
        
         \multirow{2}{*}{\small total loss  }
         & \small 100 & single-level & 67.9\\
        & \small \baseline{100} & \baseline{multi-level} & \baseline{68.3}\\[1ex]

         \multirow{2}{*}{\small total loss  }
         & \small 600 & single-level & 73.4\\
         & \small \baseline{600} & \baseline{multi-level} & \baseline{73.6}\\
    \end{tabular}
        \vspace{0.5em}
    \caption{{\bf Multi-level noise schedule}. Linear evaluation on single-level noise and multi-level noise. With a multi-level noise schedule, we typically get larger performance gain than fixed noise. }
  \label{tb:add_multi_noise}
  
\end{table}

\begin{table}[htb]
    \centering
    \begin{tabular}{l l l c }
        \bf\small Method & \bf\small Arch. & \bf\small Epochs & \bf\small Top-1\\
        \toprule
        
        \small SIMMIM~\cite{xie2022simmim} & ViT-B/16 & 100 & 56.7\\[1ex]
        \multirow{3}{*}{\small MAE~\cite{he2022masked}} & ViT-B/16 & 100 & 54.8\\
        & ViT-B/16 & 300 & 61.5\\
        & ViT-B/16 & 1600 & 67.8\\[1ex]
        \small RC-MAE~\cite{lee2022exploring} & ViT-B/16 & 1600 & 68.4\\[1ex]
        \multirow{2}{*}{\small CAE~\cite{chen2023context}} & ViT-B/16 & 300 & 64.2\\
        & ViT-B/16 & 1600 & 70.4\\[1ex]
        \multirow{2}{*}{\small SDAE~\cite{liang2021training}} & ViT-B/16 & 100 & 60.3\\
        & ViT-B/16 & 300 & 64.9\\[1ex]
         
        \multirow{2}{*}{\small IJEPA~\cite{assran2023selfsupervised}} & ViT-B/16 & 100 & 66.8*\\
        & ViT-B/16 & 600 & 72.1*\\[1ex]
        \multirow{2}{*}{\small \bf{Ours}} & ViT-B/16 & 100 & \baseline{68.3} \\
        & ViT-B/16 & 600 & \baseline{73.6} \\
        \midrule
        
        
    \end{tabular}
    \vspace{0.4em}
    \caption{{\bf ImageNet}. Linear evaluation on ImageNet-1K. Our method leads to consistent linear probing improvement compared with other methods, resulting in +1.5\% improvement on both 100 and 600 epochs settings compared with I-JEPA. * represents our reproduced results. }
    \label{tb:lineareval}
\end{table}

\section{Ablation study}
\label{sec:ablation}
In this section, we conduct an ablation study to examine each component in our N-JEPA and evaluate its effectiveness. To demonstrate this, we also assess various design options using ViT-B architecture for the mask tokens and predictor parameters. Then, we evaluate the linear
probing performance on IN-1K using only 1\% and 10\% of the available labels. We adopt a lightweight training setting for efficient evaluation: training 100 epochs with ViT-B/16, batch size 128, and 50 test epochs with batch size 1024.\\

\textbf{Noise schedule and mask token:} Table~\ref{tab:Noise ablation} details the performance of different noise schedule choices. Multi-level noise schedule serves as feature augmentations which preforms better. In table~\ref{tab:Mask token}, unshared mask tokens mean that we do not share the same mask tokens between two predictor networks.

\begin{table*}[h]
\centering
\subfloat[\textbf{Noise ablation.}  The Top.1 accuracy of different noise choices for 100 training epochs. Our multi-level noise schedule performs the best.
\label{tab:Noise ablation}]{
\centering
\begin{minipage}{0.45\linewidth}
\begin{center}
\scriptsize
\tablestyle{4pt}{1.0}
\begin{tabular}{c|cccc}
 Method & fixed noise & multi-level noise & \bf Top.1  \\ 
\shline
 Baseline          &            &    & 66.8 \\
Total loss  & \checkmark &     & 67.9 \\
Total loss  &   & \checkmark    &\baseline{\textbf{68.3}} \\        
\end{tabular}
\end{center}
\end{minipage}
}
\hspace{1em}
\subfloat[\textbf{Mask token.} Shared mask tokens between two predictors VS. Unshared mask tokens. \label{tab:Mask token}]{
\centering
\begin{minipage}{0.45\linewidth}
\begin{center}
\scriptsize
\tablestyle{4pt}{1.0}
\begin{tabular}{c|cc}
Method & Epochs   & \bf Top.1 \\ 
\shline
Total loss (with shared)    & 100 & 66.5  \\
Total loss (w/o shared)   & \baseline{100} &\baseline{\bf 67.9}  \\

\end{tabular}
\end{center}
\end{minipage}
}
\vspace{0.5em}
\caption{\textbf{Ablation studies on noise schedule and mask token.}}
\label{tab:ablations_2} 
\end{table*}

\begin{table*}[h]
\centering
\subfloat[\textbf{Predictor network parameters.}  Shared predictor network parameters  VS. Unshared predictor network parameters.
\label{tab:Predictor network parameters}]{
\centering
\begin{minipage}{0.45\linewidth}
\begin{center}
\scriptsize
\tablestyle{5pt}{1.2}
\begin{tabular}{c|cc}
Method & Epochs   & \bf Top.1 \\ 
\shline
Total loss (with shared)    & 100 & 67.3  \\
Total loss (w/o shared)   & \baseline{100} &\baseline{\bf 67.9} \\

\end{tabular}
\end{center}
\end{minipage}
}
\hspace{1em}
\subfloat[\textbf{\bf 1\% and 10\% labels of IN-1K.} Semi-supervised evaluation on ImageNet-1K using only 1\% and 10\% of the available labels for 100 training epochs.\label{tb: different_ratio_IN-1k}]{
\centering
\begin{minipage}{0.45\linewidth}
\begin{center}
\scriptsize
\tablestyle{5pt}{1.2}
\begin{tabular}{c|c c c }
        \bf\small Methods  &  \bf\small last layers &  \bf\small  1\% labels & \bf\small 10\% labels \\
        \shline
        
         \bf \small baseline & \small 1  & \small 54.3 & \small 58.5 \\
        
        \bf \small Total loss & \small 1 & \small \baseline{\bf 57.6} & \small \baseline{\bf 62.8}  \\
        \bf \small baseline & \small 4  & \small 57.5 & \small 63.0 \\
        
        \bf \small Total loss & \small 4 & \small \baseline{\bf60.0} & \small \baseline{\bf66.1}  \\
                
    \end{tabular}
\end{center}
\end{minipage}
}
\caption{\textbf{Ablation studies on predictor network parameters and various ratio of the available labels.} }
\label{tab:shared_unshared} 
\end{table*}
\textbf{Predictor Parameters:} To further seek the influence of shared predictor parameters or unshared parameters, in table~\ref{tab:Predictor network parameters}, we observe that unshared predictor parameters have slightly better performance than shared parameters.\\

\textbf{Low-Shot ImageNet-1K:}
To evaluate our model on the low-shot task, we use 1\% and 10\% of the available ImageNet labels and adapt the evaluation protocol of iBOT~\cite{zhou2021ibot}. SimCLRv2~\cite{chen2020big} found that keeping the first layer of the projection head can improve accuracy, especially under the low-shot setting. Therefore, We fine-tune the pre-trained model from the first layer of the projection head. We freeze the encoder and return the following representations: 1) the [cls] token representation of the last layer and 2) the concatenation of the last four layers of the [cls] token. We fine-tune our ViT-B models for 50 epochs on ImageNet-1\% and ImageNet-10\% with the SGD optimizer and a cosine learning rate scheduler. Our batch size is 1024.
Table~\ref{tb: different_ratio_IN-1k} shows performance on the 1\% and 10\% ImageNet benchmark. Compared with I-JEPA, our method significantly boosts the top.1 accuracy for all settings.(1\% IN-1K + 2.5\%, 10\% IN-1K + 3.1\% when using the last four layers.)

\section{Conclusion}
\label{sec:conclusion}
In this work, we introduce diffusion noise to Joint-Embedding Predictive Architecture (JEPA), namely N-JEPA, to learn more robust representations for SSL models. By injecting diffusion noise into the position embeddings of mask blocks, we ingeniously combine diffusion noise with the MIM method. Our work is a step in exploring the combination of the diffusion model and self-supervised methods. We hope our study will rekindle interest in the unified vision pretraining paradigm for recognition and generation. We will leave this extension for future work.

\clearpage
{
    \small
    \bibliographystyle{ieeenat_fullname}
    \bibliography{main}
}


\clearpage

\appendix
\section{Appendix }

\subsection{Discussion}
In our paper, we do not use larger ViT models such as ViT-L/16 and ViT-H/14 due to the limited computational resources. Consequently, our comparisons are limited to the I-JEPA framework using the ViT-B/16 model. However, we believe that our method will show consistent performance gains with larger model pretraining. Additionally,  although we aim at learning high-semantic and robust representations to enhance the performance of SSL, the capacity for generation tasks is still unexplored, we will leave it for future investigation. 

\subsection{Implementation Details}
\label{sec:implementation details}
\begin{figure*}[ht!]
\begin{minipage}[b]{.6\linewidth}
\centering
\includegraphics[width=1.0\linewidth]{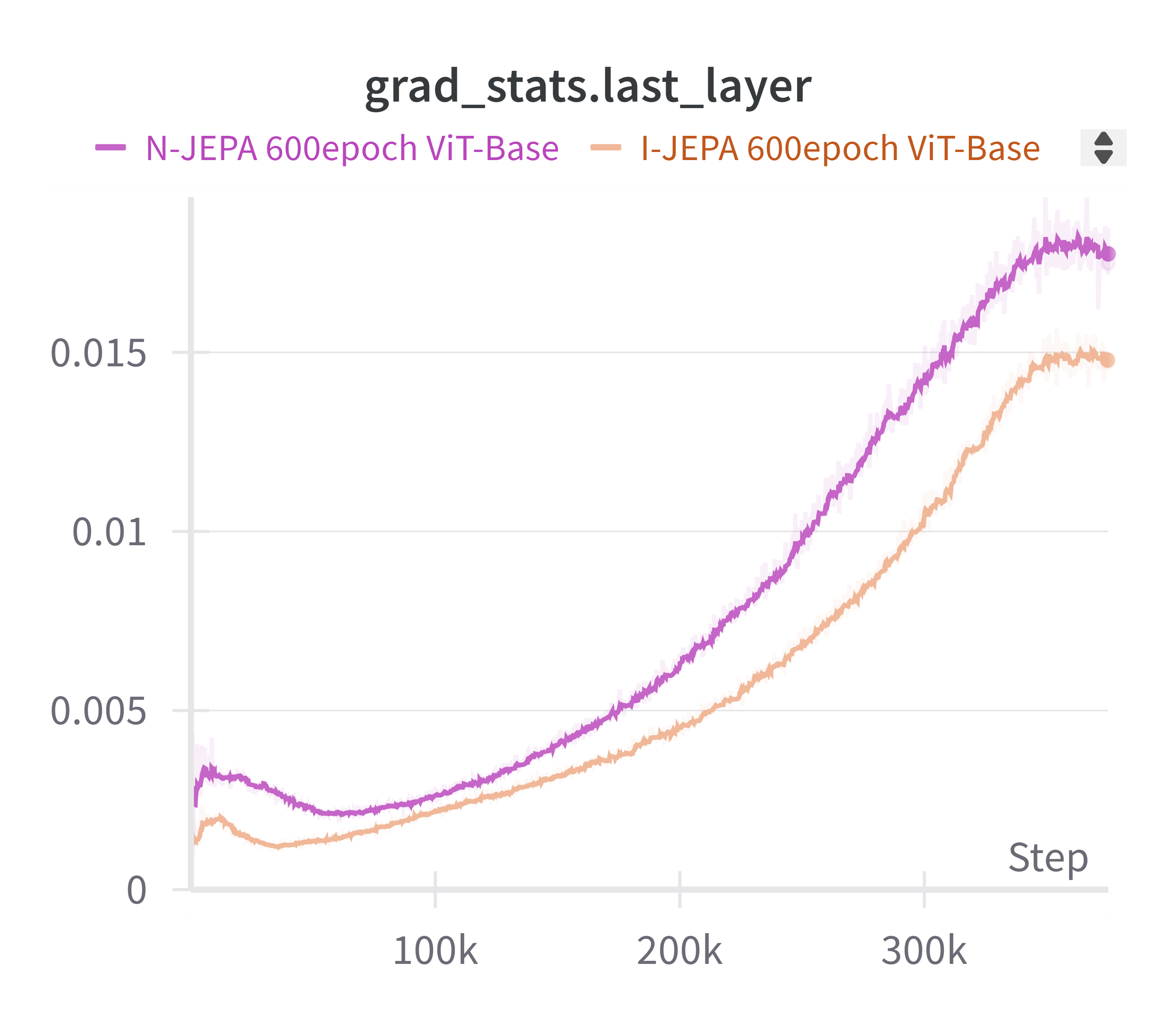}
\caption{\small \textbf{The gradient statistics of the last layer }.  }
\label{fig:grad_stats}
\end{minipage}%
\begin{minipage}[b]{.35\linewidth}
\centering
\begin{tabular}{y{102}|y{56}}
config & value \\
\shline
optimizer & AdamW \\
epochs & $600$ \\
learning rate & $3e^{-4}$ \\
weight decay &  $(0.04, 0.4)$ \\
batch size & $1024$ \\
learning rate schedule & cosine decay \\
warmup epochs & 40\\
encoder arch. & ViT-B  \\
predicted targets & 4 \\
predictor depth & 12  \\
predictor attention heads & 16 \\ 
predictor embedding dim. & 384 \\
$P_{mean}$, $P_{std}$ & -1.2 / 1.2 \\
$\sigma_{data}$  & $0.5$ \\

\end{tabular}
\vspace{-.1em}
\caption{\small\textbf{Pretraining setting for downstream tasks (ViT-B)}. All models trained for  $600$ epochs.}
\label{tab:downstream_large} \vspace{-.2em}
\end{minipage}
\end{figure*}

\begin{algorithm}[htb]
  \caption{N-JEPA pseudo-code}\label{alg:fp}
  \small
  \begin{algorithmic}[1]
    \vspace{.04in}
    \State \textbf{Input:} num iterations $K$, image dist $D$, hyper-parameter $\sigma_{data}$, \\ encoder $f_\theta$, target-encoder $f_{\bar{\theta}}$, predictor-context $g_{\phi_c}$, predictor-noise $g_{\phi_n}$, scalar $q$\\
    masked position embeddings - $\psi_{s_c}$, $\psi_{s_N}$ for predictor 1 and predictor 2. Num of mask blocks $L$
    \vspace{.04in}
    \State \textbf{Initialize:} $\bar{\theta} = \theta$ 
    
    \vspace{.04in}
    \For {$i=1,2,...,K$}
        \State \codecomment{\# sample image mini-batch, apply mask, and encode}
        \State $I_x \sim D$ 
        \State $p \sim \text{patchify}(I_x)$
        \State $x,y \leftarrow \text{student\_mask}(p),\text{teacher}(p)$ 
        \State $z_x, z_y \leftarrow f_{\theta}(x), f_{\bar{\theta}}(y)$
        \State \codecomment{\# apply \ N-JEPA, add EDM noise}

        \State $n \sim \mathcal{N}(0, \sigma_{data}^2 I)$
        \For {$j=1,2,...,L$}
        \State $\psi_{s_N}$ =  $\psi_{s_c} + n_j$
        \EndFor
        \State \codecomment{\# predict targets and compute smooth-$L1$ loss and MSE loss.}
        \State $\hat {z}_{y_c} \leftarrow g_{\phi_c}(f_{\theta}(x),\psi_{s_c}$), $\hat {z}_{y_N} \leftarrow g_{\phi_{n}}(f_{\theta}(x),\psi_{s_N}$)

        \State $\text{loss} \leftarrow ||\hat{z}_{y_c} - {z_y}\text{\scriptsize .detach()}||_2^2 + \lambda_{1}||\hat{z}_{y_N} - {z_y}\text{\scriptsize .detach()}||_2^2 + \lambda_{2}||\hat{z}_{y_N} - \hat{z}_{y_c}||_2^2 $
                \State \codecomment{\# perform sgd step and update ${\bar{\theta}}$ via ema}

        \State $\text{sgd\_step}(\text{loss}; \{\theta,\phi_{s_N},\phi_{s_c}  \})$ 

        \State $\bar{\theta} = q{\bar{\theta}} + (1-q)\theta.\text{\scriptsize detach()}$ 
    \EndFor
    \vspace{.04in}
    
  \end{algorithmic}
  
\end{algorithm}

 \end{document}